\documentclass[letterpaper, 10 pt, conference]{ieeeconf}  

\IEEEoverridecommandlockouts                              

\usepackage{makecell}
\usepackage{subfig}
\usepackage{color}
\usepackage{bm}
\usepackage{pifont}

\newcommand*\Update{\color{black}}
\newcommand*\Done{\color{black}}
\title{\LARGE \bf
 CrossMap Transformer: A Crossmodal Masked Path Transformer Using Double Back-Translation for Vision-and-Language Navigation
}

\author{Aly Magassouba$^{1}$, Komei Sugiura$^{2}$, and Hisashi Kawai$^{1}$
\thanks{$^{1}$The authors are with the National Institute of Information and Communications
Technology,
3-5 Hikaridai, Seika, Soraku, Kyoto 619-0289, Japan.
        {\tt\small firstname.lastname@nict.go.jp}}%
\thanks{$^{2}$The author is with Keio University, 3-14-1 Hiyoshi, Kohoku, Yokohama, Kanagawa 223-8522, Japan.
         {\tt\small firstname.lastname@keio.jp}}
}

\usepackage{mymacros}
\usepackage{times} 
\usepackage{graphicx}

\let\labelindent\relax
\usepackage{enumitem}
\usepackage{diagbox}
\begin{document}

\maketitle
\thispagestyle{empty}
\pagestyle{empty}


 \begin{abstract}
 Navigation guided by natural language instructions is particularly suitable for Domestic Service Robots  that interact naturally with users.  This task involves the prediction of a sequence of actions that leads to a specified destination given a natural language navigation instruction. The task thus requires the understanding of instructions, such as ``Walk out of the bathroom and wait on the stairs that are on the right''. Visual and Language Navigation remains challenging, notably because it requires the exploration of the environment and the accurate following  of a path specified by the instructions to model the relationship between language and vision. To address this, we propose the CrossMap Transformer network, which encodes the linguistic and visual features to sequentially generate a path. The CrossMap transformer is tied to a Transformer-based speaker that generates navigation instructions. The two networks share common latent features, for mutual enhancement through a double back translation model: Generated paths are translated into instructions while generated instructions are translated into paths. The experimental results show the benefits of our approach in terms of instruction understanding and instruction generation.
 \end{abstract}


\section{Introduction
\label{intro} 
}
Domestic service robots (DSRs) are promising solutions for the support of older adults and disabled people. Efforts are increasingly being made to standardize DSRs to provide various support functions \cite{iocchi2015robocup}.  Among these functions, the ability to navigate in an indoor environment is crucial as it is a pre-requisite to many daily life tasks such as fetching a glass of water from the kitchen. However, in the case of most DSRs, the ability to interact through language, while being user-friendly for the non-expert user, is limited by the complexity of understanding natural language.

In this context, we focus on understanding natural language instructions for indoor navigation. This task involves predicting a sequence of actions to reach a goal destination from instructions such as ``{\it Go down stairs. At the bottom of the stairs walk through the living room and to the right into the bathroom. Stop at the sink.}'' Such a task presents several challenges related to the ambiguity of the instructions because the many-to-many nature of mapping between language and the real world makes it difficult to accurately predict user intention. In particular, unlike  the understanding of manipulation instruction  \cite{magassouba2018multimodal, magassouba2019understanding} based on a single environment state, this task requires the language to be mapped to changing states as the DSR moves towards the destination. Furthermore, navigation instructions are generally longer than manipulation instructions, which increases the complexity of the task.  

The task has recently been formalized as visual-and-language navigation \cite{anderson2018vision} (VLN), and  many approaches based on data-driven methods\cite{fried2018speaker, tan2019learning} have been proposed. Classically, these approaches exploit a recurrent neural network to infer a sequence of actions leading to the desired destination. Although, these approaches have shown promise, their level of accuracy remains far from that of a person \cite{anderson2018vision}.

\begin{figure}[tp]
   \centering
      \includegraphics[width=0.95\columnwidth]{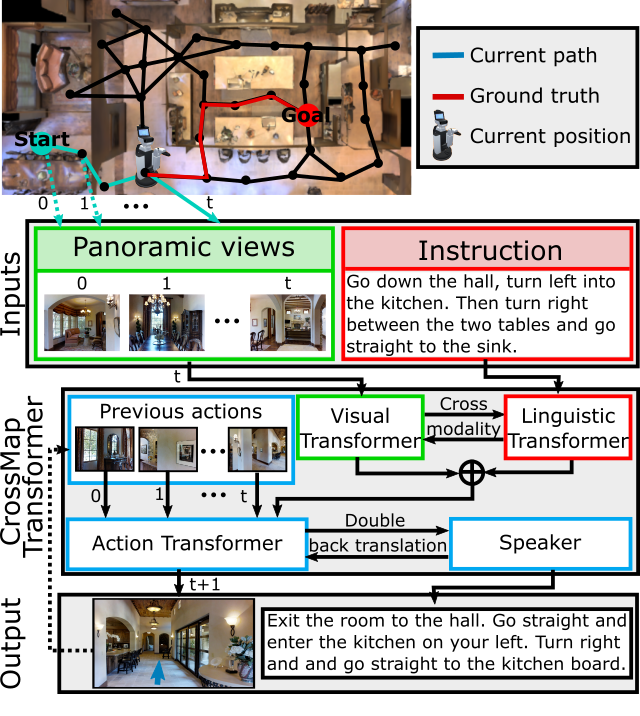}
      \caption{\small Our approach, the CrossMap Transformer is used to predict the sequence of actions to navigate to a goal destination given an instruction.  }
      \label{fig:architecture}
 \end{figure}

To narrow this accuracy gap, we propose the Cross-modal Masked Path (CrossMap) Transformer.  Motivated by the recent development of transformer networks \cite{vaswani2017attention} for language modeling \cite{devlin2018bert,dong2019unified}, the CrossMap Transformer encodes linguistic and environment state features to sequentially generate actions similarly to recurrent network approaches. Our approach uses feature masking to better model the relationship between the instruction and environment features. 

Additionally, we use a double back translation (DBT) approach. Unlike \cite{fried2018speaker} where a speaker network is trained separately to generate instructions, the DBT consists in mutually training the speaker network, and the CrossMap Transformer, by using common latent features.

The present work makes the following key contributions to the literature:
\begin{itemize}
\item We propose the CrossMap Transformer, which sequentially generates actions to reach an instructed destination from linguistic and environment state features. We explain the method in Section \ref{sec:method}.
\item We propose a double back-translation to improve the mapping between linguistic and action features as explained in Section \ref{sec:method}.
\item We apply the CrossMap Transformer on the standard Room-to-Room(R2R) dataset \cite{anderson2018vision}. Our approach achieves results comparable to those of recently proposed state-of-the-art recurrent neural network methods. We present the experimental validation in Section \ref{sec:exp}.
\end{itemize}


\section{Related Work
\label{related}
}

Robot navigation \cite{xia2020interactive} and path planning from natural language instructions have been widely investigated in the field of robotics \cite{kollar2010toward, tellex2011understanding, kuo2020deep}.
Such a task was recently formalized adopting data-driven  methods \cite{anderson2018vision}  with the release of the R2R dataset. 
In this setup, the VLN task \cite{mogadala2019trends} is addressed using Long Short Term Memory (LSTM) networks structured in an encoder-decoder framework. An instruction is encoded first and then decoded as a sequence of actions using the current environment states. 
Initially, the VLN method uses low-level action spaces, where each motion (e.g., left, right or forward motion) of the robot is predicted. 
The use of a panoramic action space \cite{fried2018speaker} has been shown to improve results as the action sequence directly moves the robot from one position to another.
However, the complexity of the VLN task is emphasized by the large gap between human performance and the performance of  neural models. Additionally, the trade-off between exploration and instruction fidelity has been emphasized in \cite{jain2019stay}  and additional evaluation metrics, such as the dynamic time warping, have been proposed.  

To overcome these limitations, several works \cite{fried2018speaker, wang2019reinforced} proposed  exhaustive exploration of the environment. Although these approaches are generally more accurate, they are not feasible in real-world environments.
Another line of work relates to data augmentation \cite{Yu_2020_CVPR_Workshops}. Indeed, the R2R dataset is relatively small and introduces several biases in the training set distribution\cite{anderson2018vision}. To mitigate these biases, the Room-for-Room (R4R) dataset was introduced in \cite{jain2019stay} by synthetically  concatenating several paths and instructions. A back-translation model, a speaker \cite{fried2018speaker}, has been introduced to generate additional instructions for unlabeled path. This approach was extended in \cite{tan2019learning} with the environmental dropout, to improve the back-translation model by synthetically generating new environments from dropped features. The exploration of auxiliary tasks \cite{zhu2020vision}, such as ensuring the consistency of the path has had positive results. Globally, pre-training on a large dataset has been proven to improve the generalization of  VLN models \cite{majumdar2020improving, hao2020towards}.

Our approach, using the CrossMap Transformer, is based on transformer networks. Although such structures have been widely used in language modeling \cite{dong2019unified, wang2020minilm, devlin2018bert}, few works have used transformers to address the VLN task. \Update In \cite{majumdar2020improving},  transformers were used to score the compatibility between paths and instructions, in the setting of the  pre-exploration of an environment. In line of works such as \cite{li-etal-2019-robust, hao2020towards}, transformers were used in pretraining the language embedding model by combining visual and linguistic features. \Done The path was nonetheless predicted from an LSTM-based encoder-decoder. Conversely, \cite{landi2019perceive} proposed to using transformers to decode the instructions and environment features as a sequence of actions. Nonetheless, there remains a gap in performance between this approach and classic LSTM-based networks.



\section{Problem Statement
\label{sec:problem}
}

\begin{figure}[tp]
   \centering
      \includegraphics[width=1\columnwidth, height=2.8cm]{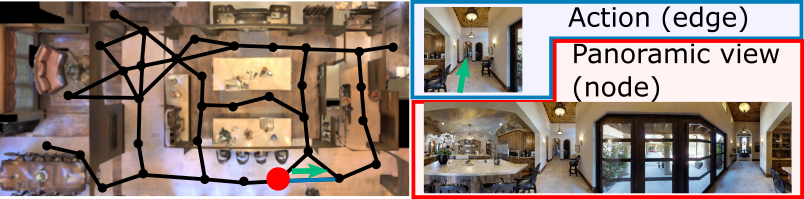}
      \caption{\small For the VLN task, each environment is given as a navigation graph where each node is a panoramic view and each edge an action to move from one node to another one.  }

   \label{fig:R2Rsample}
 \end{figure}

\begin{figure*}[tp]
  \centering
   \subfloat{\includegraphics[width=0.95\textwidth]{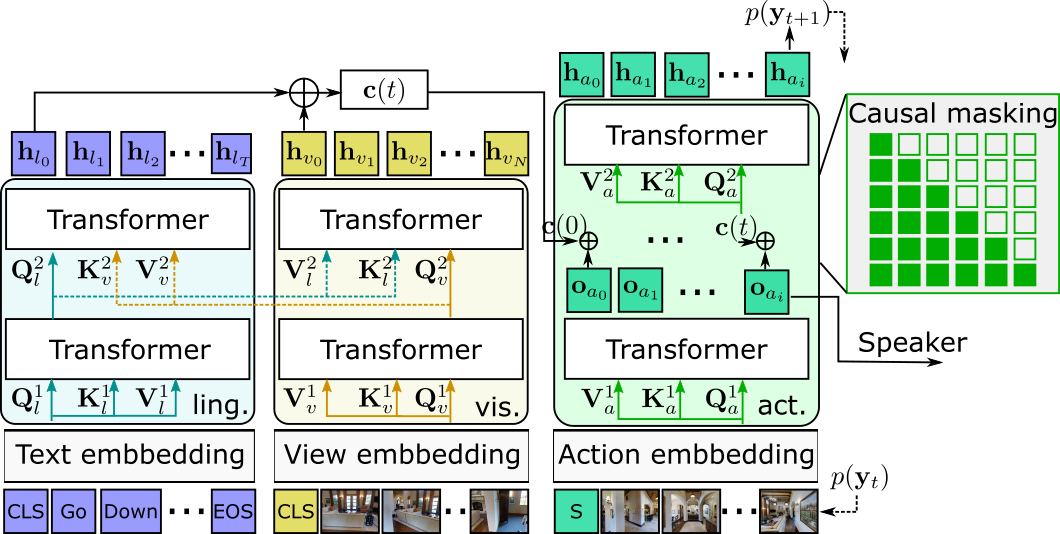}}
    \caption{\small  The CrossMap Transformer (CMT) model endows linguistic, visual and action transformers to predict the sequence of actions given an instruction and a causal masking structure.} \label{fig:cmt} 
\end{figure*}

\subsection{Task background}
We aim to solve the VLN task in real-world home environments as described in \cite{anderson2018vision}. More specifically, given a natural navigation instruction, this task involves predicting a sequence of actions that leads to a goal destination. This task requires the understanding of  instructions such as ``Walk out of the bathroom and wait on the stairs that are on the right'' and ``Go down stairs. At the bottom of the stairs walk through the living room and to the right into the bathroom. Stop at the sink.'' Such a task is particularly relevant to DSRs, as it complements manipulation tasks \cite{magassouba2019understanding, magassouba2018multimodal} for communicative robots that interact naturally with users.
Nonetheless, the VLN task is challenging because it requires natural and long sentences to be addressed, where the relationships among individual words and with the current environment state should be modeled. 

Additionally, this task requires a trade-off between exploration of the environment and fidelity to the instruction. Indeed, it has been shown that methods based on exploration outperform methods that learn only from the ground truth paths \cite{anderson2018vision}. This is explained by the difference in the distributions of the training and test data when learning only using ground truth methods. Meanwhile, although exploration yields better results, such an approach induces discrepancies between the instruction and generated path \cite{jain2019stay} which limits the modeling between the instruction and physical world. 

Hence, many studies \cite{fried2018speaker, majumdar2020improving} have propose exhaustively exploring the environment and selecting the best path to reach the specified destination. However, considering a physical deployment on DSRs, such methods are cumbersome and time consuming as an exhaustive map should be built and explored before performing the required task. As so, the work, we consider single run approaches, where only one path is generated on-the-fly.

\subsection{Task Description}
The task environments of different size are in real home scenes as depicted in Fig. \ref{fig:R2Rsample}. Each scene is discretized into connected nodes that build different paths. A node corresponds to a 360 degree panoramic image. 

Similarly to many other works \cite{fried2018speaker, hao2020towards}, we consider that the VLN is performed through adopting a panoramic action space instead of atomic actions \cite{anderson2018vision} (i.e., atomic motion). 
Hence, the navigation task can be represented as a graph where each node represents a waypoint in the scene (see Fig. \ref{fig:R2Rsample}). Each edge of this graph is an action to move from one node to another. A path is then a sequence of nodes from an initial node to a destination node.

Unlike the aforementioned studies \cite{fried2018speaker, majumdar2020improving} that addressed this task through full knowledge of the navigation graph, we consider a single-run setup where only the current node and adjacent edges are known. 

In this setup, the VLN task is sequential and for every time step $t$ the following inputs and outputs are expected:
\begin{itemize}
    \item[$\bullet$]{\bf Inputs} $(t=0)$: A navigation instruction as a sentence.
    \item[$\bullet$]{\bf Inputs} $(t \geq 0)$: The current node (waypoint), the adjacent edges, and a panoramic image taken at the node (waypoint).
      \item[$\bullet$]{\bf Output} $(t \geq 0)$: Next action (edge) among the adjacent edges.  
\end{itemize}

Four evaluation metrics are considered in this study, that is, the Success Rate (SR)  which is the rate of successfully generating a path (arriving within 3 meters of the desired destination),  the navigation error (NE), which is the mean distance of arrival from the desired destination, the Success weighted by Path Length (SPL), which is the ratio of successful predictions normalized by the path length, and the oracle success rate (OSR), which is the rate of generated paths that cross within 3 meters of the desired destination.


\section{Proposed Method
\label{sec:method}
}
\subsection{Novelty}
As explained in the previous section, there is a gap in performance between state-of-the-art methods using the LSTM architecture \cite{tan2019learning} and those using transformers \cite{landi2019perceive}. Indeed, transformers are generally used for feature modeling \cite{devlin2018bert, chen2020uniter},  achieving state-of-the-art results. \Update Text generation has also been addressed using transformers \cite{dong2019unified, radford2018improving} and causal masking \cite{radford2019language}. In this line of works, each predicted token attends only tokens at previous positions. Nonetheless very few transformer architectures are optimized for the VLN task. Yet transformers can be trained faster than LSTMs \cite{vaswani2017attention} for long sequences and can take advantages of the recently released pretrained models on generic tasks using linguistic and visual modalities \cite{chen2020uniter, zhou2020unified}.
 
Inspired by this line of work, we propose a transformer architecture for VLN, the CrossMap Transformer (CMT) illustrated in Fig. \ref{fig:cmt}.  our approach predicts a sequence of actions that reaches the specified goal via causal masking  and at the same time learns the relationship between the instruction and the corresponding path. Furthermore, we introduce a transformer-based speaker, CrossMap Speaker (CMS) to improve the generalization ability of our approach through a double back-translation (DBT) model. We advocate that a better model can be obtained when mutually training the CMS and CMT networks with common latent features. 
\Done

The CMT has the following characteristics:
\begin{itemize}
\item The CMT combines visual and linguistic modalities to predict a path through a sequence of actions.
\item The CMT adopts cross-modal path masking to model the relationship between the navigation path and instruction.
\item The CMT and CMS are mutually trained through DBT, which enhances both networks.
\end{itemize}

\subsection{CrossMap Transformer Architecture}

\subsubsection{Network Inputs}
Let us consider an instruction $i$, such that at each time step $t$ of the sequence, the set of inputs of the network is defined as:
\begin{equation}\label{eq:input}
    \bm{ x}^i(t)= \{\bm{ x}_l^i, \bm{ x}_c^i(t), \bm{ x}_a^i(t)\},
\end{equation}
where $\bm{ x}_l^i$  and $\bm{ x}_a^i(t)$ respectively denote the linguistic and previous action inputs and $\bm{ x}_c(t)$ is the current navigation node. In the following, the index $i$ is omitted for simplicity.

In detail, $\bm{ x}_l$ is the embedded instruction. The current navigation node is given as a set of N image views  so that
\begin{equation}\label{eq:pan}
    \bm{ x}_c(t)=\{I_0(t), I_1(t),\dots, I_n(t), \dots, I_{N}(t)\},
\end{equation}
where each image $I_n(t)$ is a portion of the input panoramic image at step $t$. In this study, we consider that $N=36$ as proposed in \cite{anderson2018vision}. 

The previous actions are given as a sequence
\begin{align}\label{eq:act}
    \bm{ x}_a(t)=\{E(0), E(1),\dots, E(t-1)\},
\end{align} 
of the $(t-1)$ steps actions performed. An action $E(t)$ is defined as a given view of the current panorama in the direction of the next navigation node. 

\subsubsection{Language Encoder}
Each instruction $\bm{ x}_l$ is initially tokenized into subwords that are then embedded as 384-dimensional vectors using MiniLM network \cite{wang2020minilm}. MiniLM is the distillated version of UniLM \cite{dong2019unified}, which is a state-of-the-art method of masked language modeling based on transformers. 
These vectors are then processed through the language encoder. The language encoder is a two-layer transformer that takes as input the tokens features and the corresponding positional encoding. The maximum instruction sequence length is set to 42, that is, 40 tokens and as the beginning-of-sentence ($<$CLS$>$) and end-of-sentence ($<$EOS$>$) tokens. Each sentence is padded according to its length with the padding token ($<$PAD$>$). 

In the language encoder, self-attention heads process the input vectors and are followed by a fully connected feed-forward network.
The output of an attention head is given by \cite{vaswani2017attention}: 
 \begin{equation}\label{eq:attention}
    \bm{A}= \bm{V} \text{softmax}\left( \frac{\bm{QK}^T}{\sqrt{d_k}} + \bm{M} \right),
\end{equation}
where $\bm{ Q}$, $\bm{ K}$ and $\bm{ V}$ are the queries, keys and values, whereas $\bm{ M}$ is a mask matrix, controlling where each token can attend. The mask $\bm{ M}$ takes a value of  ${0, -\infty}$ to allow or prevent attention. 
Within this transformer,  self-attention and  linguistic cross-modal attention are applied  as defined in \eqref{eq:attention} by also considering the current environment state through the $\bm{ x}_c(t)$ to condition the language input. In the self-attention configuration, $\bm{ M}$ corresponds to a linguistic mask matrix while $\bm{ Q}$, $\bm{ K}$ and $\bm{ V}$ are projections of the transformer hidden vector. For the linguistic cross-modal attention, the queries $\bm{ Q}$ and values $\bm{ V}$ are projections of the visual features $\bm{ x}_c(t)$. 
The $<$CLS$>$ is used as the current representation of the instruction and is output as a vector $\bm{ h}_{l_0}(t)$  of dimension 1$\times$384.

\subsubsection{Visual Encoder}
Each image is encoded as the concatenation of the semantic view provided in \cite{chang2017matterport3d} and the ResNet-152 \cite{he2016identity}  features as provided in \cite{anderson2018vision}. The intuition behind this approach is to use different granularity of features. Indeed, ResNet-152 provides low-level features while the semantic features provides to high-level features.

A positional feature is also input to the visual encoder by concatenating each visual feature with the set of features $[ \cos(\theta_n), \sin(\theta_n), \cos(\phi_n), \sin(\phi_n)]$ that encodes the relative azimuth $\theta_n$ and elevation $\phi_n$ of a view $I_n$ with respect to the current robot orientation.

Similarly to the linguistic encoder,  cross-modal attention is used in the visual encoder. Indeed, following the self-attentive transformer layer,  each attention head is conditioned by linguistic features obtained from the linguistic encoder. A visual representation at the step $t$ is obtained using $<$CLS$>$ token. This token is processed and output as a vector $\bm{ h}_{v_0}(t)$ of dimension 1$\times$384.

\begin{figure}[tp]
  \centering
   \subfloat{\includegraphics[width=0.95\columnwidth]{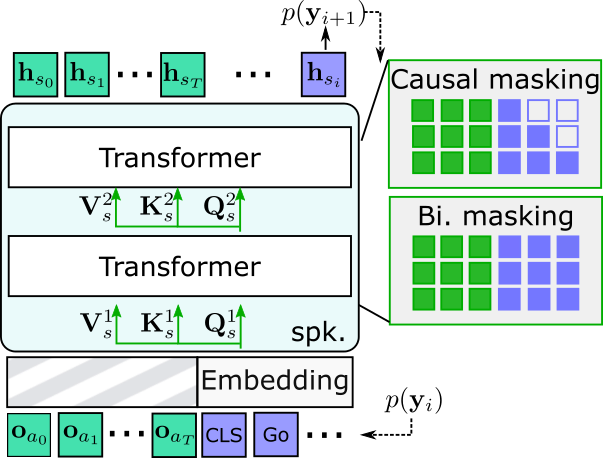}}
    \caption{\small CrossMap Speaker (CMS) configuration that combines the sequences of latent actions features from the CrossMap Transformer (CMT) and the instructions. The CMS uses both causal and bidirectional masking.} \label{fig:spk} 
\end{figure}

\subsubsection{Action Decoder}

The action decoder predicts the likelihood of all candidate actions given the linguistic and visual features at step $t$ and the previous sequence of actions. 
Each action is embedded similarly for each panoramic view, that is by concatenating the high-level semantic features and low-level ResNet-152 features. Additionally, two types of positional encoding are used for the action decoder. The first positional encoding corresponds classically  to  the position of the action in the sequence.
For the second type of positional encoding, each action is concatenated with the set of features $[\cos(\theta_m), \sin(\theta_m), \cos(\phi_m), \sin(\phi_m), \rho_m]$ that encodes the relative azimuth $\theta_m$ and elevation $\phi_m$ and distance $\rho_m$ to  an adjacent node $m$ with respect to the current robot pose. For the stop action, the azimuth and elevation angles as well as the distance are set to zero.

These inputs are processed by the action decoder. The first layer outputs $\{ \bm{ o}_a(0), \dots \bm{ o}_a(t)\}$,  which is then concatenated with the corresponding context feature $\bm{C}(t) ~=~ \bm{h}_{l_0}(t) \oplus~\bm{h}_{v_0}(t)$.  The features $\{ \bm{o}_a(0) \oplus \bm{C}(0), \dots , \bm{o}_a(t) \oplus \bm{C}(t) \}$ are then processed by the second layer of the transformer. It is noted that the intermediate outputs $\{ \bm{ o}_a(0), \dots , \bm{ o}_a(t)\}$ are also used for the CMS network. Such an architecture allows training both the CMT and CMS.
For the action decoder, a causal masking is used, that is $\bm{ M}$ corresponds to a mask where each embedded feature attends only the previous features of the sequence.
The last output $\bm{ h}_a(t)$ of the transformer is then used to compute the likelihood $p(\bm{ y})(t + 1)$ given as
 \begin{equation}\label{eq:out}
    p(\bm{ y})(t + 1) = \bm{ h}_a(t) \bm{ y}(t+1),
\end{equation}
where $\bm{ y}$ corresponds to the set of candidate actions from the current robot position.

\subsubsection{Loss function}
To predict the likelihood $p(\bm{ y}(t))$ of an action, the loss function $\mathcal{L}$  as the cross-entropy function is expressed as:
\begin{align} \label{equ:J}
    \mathcal{L} &= -\sum_n \sum_{m} y^{*}_{nm} \log p(y_{nm}),
\end{align}
where $y^{*}_{nm}$ denotes the label given to the $m$-th dimension of the $n$-th sample, and $y_{nm}$ denotes its prediction. As performed in \cite{anderson2018vision}, the next action $\bm{ x}_a(t + 1)$ in the sequence is sampled from $p(\bm{ y}(t))$ to allow the agent to explore the environment.

\subsubsection{Cross-modal Path Masking}
A novelty of the CMT is the cross-modal path masking. In addition to the loss function describe above, the CMT learns the relationship between the ground-truth paths and the corresponding instruction. Similarly to Natural Language Processing (NLP) methods such as BERT \cite{devlin2018bert}, a ground-truth sequence of action is randomly masked, and predicted by the network. Given a ground-truth sequence of action  $\{ \bm{ y}(0), \dots \bm{ y}(T-1), \bm{ y}(T)\}$ of size $T$, and a masked position $m$, $m \le T$, the CMT computes  $p( \bm{ y}(m)|\bm{ y}(m - 1), \dots , \bm{ y}(0))$ through a causal mask and the same transformer architecture described previously. This approach is used in the pre-training phase.

\subsubsection{Speaker}
In addition to the CMT, we introduce the CMS (see Fig. \ref{fig:spk}) network that is a two-layer transformer generating a sentence from a sequence of actions. The CMS uses a similar architecture as visual and language transformers such as VLP \cite{zhou2020unified}.  The inputs of the CMS are the full sequence of latent action features $\bm{ o}_a(t)$  that are computed by the CMT. The CMS generates the sequence of words using alternatively causal masking to generate the sequence of words and bi-directional masking to learn the relationship between the instruction and the sequence of actions.

\subsubsection{Double back-translation}
Such an architecture where the CMT and CMS share the same features representation allows $\bm{ o}_a(t)$ to be concurrently processed into a sequence of actions and translated into an instruction. This approach mutually enhances the CMT and CMS network, differently to the method proposed in \cite{fried2018speaker} where the speaker is trained separately. In addition, more classically,  we used a second translation by using the sentences generated by the CMS to train the CMT.
We define these two levels of translation as the double back translation (DBT). 

\Update
To perform the DBT, the CMS minimizes the cross-entropy loss between the generated sentence and the original instruction. In this configuration,  each input sample $\bm{ o}_a(t)$, of a full sequence generated by the CMT, is used to train the CMS only if the instructed destination is successfully reached. A mask is used to discard the samples  that did not reach the correct destination. 

In a second phase the CMS generates sentences from latent features $\bm{ o}_a(t)$ obtained from ground-truth paths. These sentences are used to train the CMT. Only sentences with a scoring metric greater than a threshold $\lambda$ are used.

During the inference phase, if the average CMS scoring metric of the validation set is greater than $\lambda$, for the next epoch, the CMS is then used to generate sentence on the unlabeled training paths. These newly labeled paths are used as augmented data to train the CMT. By filtering both the CMT and CMS,  we limit the impact of noisy generated features as analyzed in \cite{zhao2021evaluation}. 

\Done

\section{Experiments
\label{sec:exp}
}

\subsection{Experimental Setup}
\begin{table}[t]
\small
\centering
\caption{\small Parameter settings and structures of the CrossMap Transformer (CMT)}\label{tab:param}
\begin{tabular}{|l|l|}
\hline
CMT & Adam $( \mbox{lr}= 5 \times 10^{-4}$, \\
 Opt. method  &$\beta_1=0.99$, $\beta_2=0.9 )$ \\
 
\hline
Nb. layers  & $2$\\
\hline
Hidden size & $348$\\
\hline
Language Model  & MiniLM \cite{wang2020minilm} \\
\hline
 Activation & ReLU  \\
\hline
Nb. heads  & $12$\\
\hline
Feed-forward size & $1534$  \\
\hline
Dropout  & $0.1$  \\
\hline
Env. Dropout  & $0.4$\\
\hline
$\lambda$ (SPICE)  & $20$\\
\hline
 Batch size & 50  \\
 \hline

\end{tabular}
\end{table}

\begin{table*}[t]
\normalsize
\caption{\small Comparison results of the VLN task for LSTM-based decoder and Transformer-based decoder under several metrics: Success Rate (SR), Navigation Error (NE), Success Path Length (SPL) and Oracle Success Rate (OSR). The results for PREVALENT* are obtained from our own implementation. }
\label{tab:results}
\centering
\begin{tabular}{l|c|cccc||cccc}
\hline
 &{\bf Decoder} &\multicolumn{4}{c||}{\bf Validation Seen } &\multicolumn{4}{c}{\bf Validation Unseen } \\
\cline{3-10}
 \multicolumn{1}{l|}{\bf Method }  &{\bf  Type}& {\bf SR$\uparrow$}& {\bf NE$\downarrow$} &{\bf SPL$\uparrow$}&{\bf OSR$\uparrow$} & {\bf SR$\uparrow$}& {\bf NE$\downarrow$} &{\bf SPL$\uparrow$}&{\bf OSR$\uparrow$}\\
\hline 
Seq2Seq \cite{anderson2018vision}& &0.39&6.01&$-$&0.53 &0.22&7.81&$-$&0.28\\
Speaker-Follower \cite{fried2018speaker} & &0.66&3.36&$-$&0.74 &0.35&6.62&$-$&0.45\\
RCM \cite{wang2019reinforced}& &0.67&3.37&$-$&0.77 &0.43&5.88&$-$&0.52\\
EnvDrop \cite{tan2019learning} &LSTM &0.62&3.99&0.59&$-$ &0.52&5.22&0.48&$-$\\
AuxRN \cite{zhu2020vision} & &0.70&3.33&0.67&{\bf 0.78} &0.55&5.28&0.50&0.62\\
PREVALENT \cite{hao2020towards} & &0.69&3.67&0.65&$-$ &0.58&4.71&{\bf 0.53}&$-$\\
PREVALENT* & &0.70&3.48&0.67& 0.77 &0.57&4.66&0.52&{\bf 0.65}\\
CMG-AAL \cite{zhang2020language}& & {\bf 0.73}& {\bf 2.74}& {\bf 0.69}&$-$ &{\bf 0.59}& {\bf 4.18}&0.51&$-$\\
\hline
 PTA \cite{landi2019perceive} & &0.66&3.35& {\bf 0.64} &0.74 &0.43&5.95&0.39&0.49\\
\Update CMT (BT-only)\Done& Transformer &\Update 0.67\Done&\Update 3.43\Done&\Update 0.59\Done&\Update0.74\Done&\Update 0.51\Done&\Update 4.89&\Update 0.40\Done&\Update 0.58\Done\\
CMT (Ours) & &{\bf 0.73}&{\bf 2.82}&0.63&{\bf 0.80} &{\bf 0.55}& {\bf 4.60}&{\bf 0.44}&{\bf 0.63}\\

\end{tabular}
\squeezeup
\end{table*}

Parameters of the CMT are summarized in Table \ref{tab:param}. 
Each of the transformers comprised two layers of 12 attention heads, with a hidden dimension of 384, while the feed-forward layers had a dimension 1,534. We applied a dropout rate of 0.1 to each layer.
In the visual encoder, we used the environmental dropout on the  visual features with a rate of 0.4.

The training procedure was divided into two phases. First, both the CMT and CMS were mutually pre-trained using ground truth paths and cross-modal path masking. In the second phase, both networks were fine-tuned following the setup given in Section \ref{sec:method}. Unlike previous works, we selected Spice \cite{anderson2016spice} as the CMS scoring metric, and set $\lambda=20$. It was shown in \cite{zhao2021evaluation}, that this metric is most consistent with human labelling in the VLN for the task.

The CMT was trained on a machine equipped with four Tesla V100 with 32 GB of memory, 768 GB RAM and an Intel Xeon 2.10 GHz processor. The results were reported after 300 epochs. With this setup, it took around 1 day  to train the CMT with a batch size of 50 samples and at learning rate of 5$\times$10$^{-4}$.

\subsection{R2R Dataset}
The R2R dataset \cite{anderson2018vision} is based on the Matterport3D \cite{chang2017matterport3d}  environments annotated with navigation instructions. The dataset contains 21,567 instructions, which are 29 words long on average, for 90 different environments. The dataset was split with 14,025 instructions in 60 different environments as the training set. Two different validation sets were considered. The first one, validation seen,  contained 1,020 instructions for the same environments as the training set. The second one, validation unseen, contained 2,349 instructions for 11 environments different from the training set environments.

Additionally, for data augmentation, we used the approx. 170,000 collected paths given in \cite{fried2018speaker}. These paths were collected in the training environments and were initially unlabeled. 

\begin{figure*}[t]
\centering
\subfloat[Ground Truth (GT): ``Enter the bedroom. Turn left and exit the bedroom through the door.  Wait by stairs''. $||$ CMS: ``Exit the bathroom and turn left. walk past the bed and exit the room. wait there''.]{\includegraphics[width=0.95\textwidth]{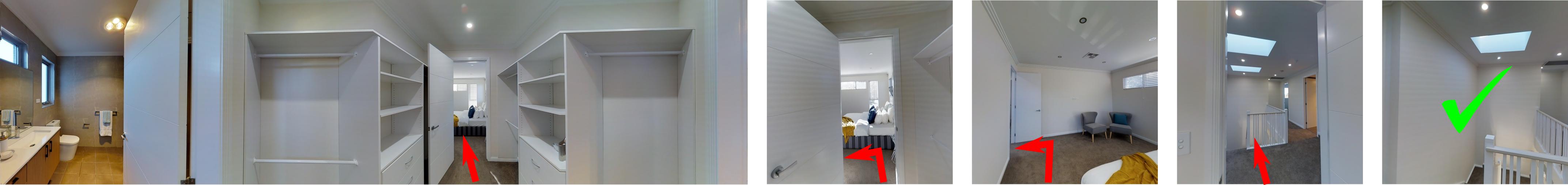}}\\  \squeezeup
\subfloat[GT: ``Go down the stairs, go slight left at the bottom and go through door, take an immediate left and enter the bathroom, stop just inside in front of the sink''. $||$  CMS: ``Go down the stairs and turn left. walk straight and enter the room on the left. stop in front of the sink''. ]{\includegraphics[width=0.95\textwidth]{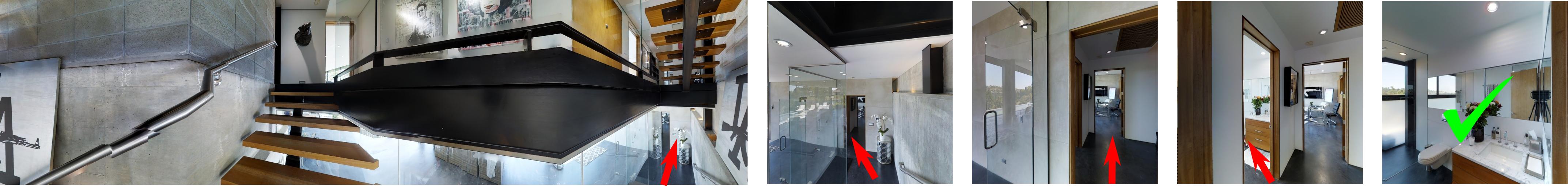}}\\ \squeezeup
\subfloat[GT: ``Take a right and walk out of the kitchen. Take a left and wait by the dining room table''. $||$  CMS: ``Walk past the kitchen and turn left. walk past the dining room table and chairs and stop''.]{\includegraphics[width=0.95\textwidth]{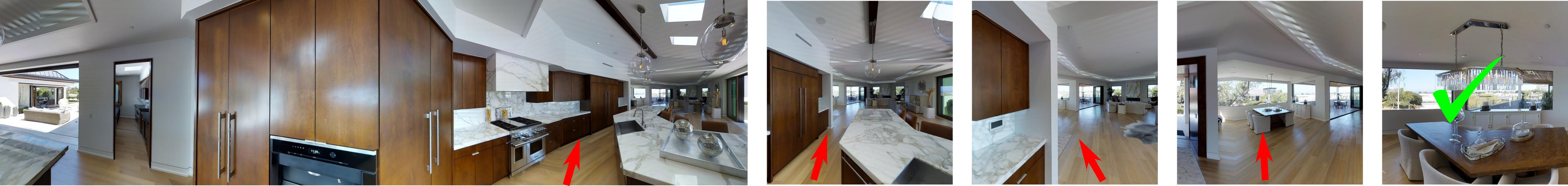}}\\ \squeezeup
\subfloat[GT: ``Head around the table and go the main area left of the long table.  Go to the middle of the room next to the ping pong table and stop''. $||$  CMS: ``Walk past the ping pong table and wait by the ping pong table''.]{\includegraphics[width=0.95\textwidth]{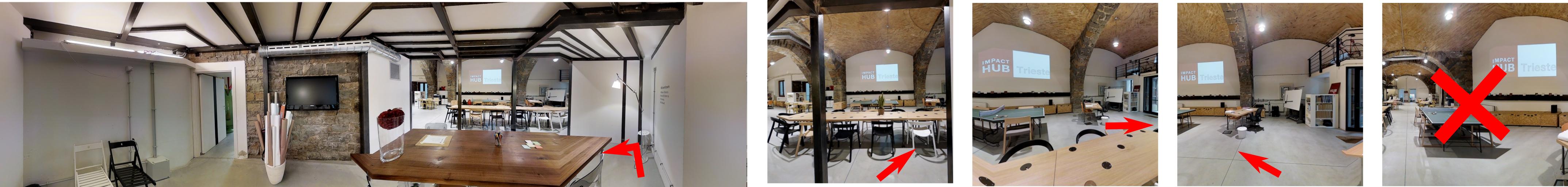}}
\caption{ \small  Qualitative results of the CMT represented as a sequence of actions from an initial panoramic image and a ground truth (GT) instruction. The generated sentences from the CMS are also given.}

\label{fig:qualitative}
\end{figure*}

\subsection{Quantitative Results}
We compared the CMT with state-of-the-art methods in terms of the metrics defined in Section \ref{sec:problem} for the two validation sets of the  R2R dataset. It is emphasized that we performed a comparison with approaches that use the same setups as in this study. More explicitly, these setups are the panoramic action space and single run setups, without pre-exploration or beam-search.

The results reported in Table \ref{tab:results} indicates that our approach obtained results, except for the SPL metric, comparable to those of the currently best performing methods, PREVALENT \cite{hao2020towards} and CMG-AAL\cite{zhang2020language}. When compared with another Transformer-based approach such as PTA \cite{landi2019perceive}, the CMT achieved better results for all metrics except the SPL  in the case of the validation seen dataset.
We hypothesize that the lower performance  in terms of  SPL metric may be related to the fact that we did not optimize the CMT for the shortest path unlike \cite{tan2019learning,fried2018speaker}. These works adopt a reinforcement learning approach, which penalizes, among other things, the number of step taken (longer paths). Nonetheless, it is worth mentioning that the SPL scores the optimal (shortest) path but fails to take into account the similarity between the reference and generated trajectory as analyzed in \cite{jain2019stay}. 

\Update
\begin{table}[t]
\normalsize
\caption{\small Effect of path masking and DBT of the CMT for the validation sets. The values in $(\cdot)$ represent the standard deviations. CMT-type1 refers to a model without DBT and path masking, while CMT-type2 is a model without DBT.
}
\label{tab:abl1}
\centering
\begin{tabular}{l|cccc}
\hline
 &\multicolumn{4}{c}{\bf Validation Seen } \\
\cline{2-5}
 \multicolumn{1}{l|}{\bf Method }  & {\bf SR$\uparrow$}& {\bf NE$\downarrow$} &{\bf SPL$\uparrow$}&{\bf OSR$\uparrow$} \\
\hline
CMT-type1  &0.59&4.00&0.51&0.70 \\
 & \Update (0.01)\Done & \Update (0.16)\Done& \Update (0.03)\Done& \Update (0.04)\Done \\
 \hline
CMT-type2 &0.64&3.71&0.53&0.71 \\
& \Update(0.02)\Done& \Update(0.33)\Done& \Update(0.03)\Done& \Update(0.03)\Done
\end{tabular} 
\end{table}   
\begin{table}
\normalsize
\label{tab:abl2}
\centering
\begin{tabular}{l|cccc}
\hline
&\multicolumn{4}{c}{\bf Validation Unseen } \\
\cline{2-5}
 \multicolumn{1}{l|}{\bf Method }  & {\bf SR$\uparrow$}& {\bf NE$\downarrow$} &{\bf SPL$\uparrow$}&{\bf OSR$\uparrow$} \\
\hline
CMT-type1  &0.48&5.44&0.35&0.56 \\
& \Update (0.02)\Done & \Update (0.13)\Done & \Update (0.02)\Done & \Update (0.03)\Done \\
\hline
CMT-type2&0.50&5.01&0.38&0.55 \\
&\Update(0.02)\Done & \Update(0.14)\Done &\Update(0.04)\Done &\Update(0.02)\Done\\
\end{tabular} 
\end{table}  
\Done

\Update
Additionally our result CMT (BT-only) that uses only back-translation as in  \ref{tab:results}), emphasizes the relevance of using a DBT approach to improve both validation seen and unseen metrics. Furthermore, this result shows that using a transformer-based decoder can be beneficial to the VLN task, as the CMT (BT-only) generalize better on the validation datasets than the Speaker-Follower or EnvDrop methods. Indeed, EnvDrop improves the Speaker-Follower for the unseen dataset, but performs worse on the unseen dataset.
\Done

To gain a better insight into the CMT, we also performed an ablation study considering the cross-modal path masking, as well as the DBT; results are given in Tables \ref{tab:abl1}. We considered two ablation conditions, CMT-type1 that is without DBT and path masking, and CMT-type2 that is without DBT. The results emphasize that both ablated parts greatly improve the performance of the CMT.

\subsection{CrossMap Speaker results}
An alternative way to assess the contributions of this study is to evaluate the performance of the CMS network. As stated in Table \ref{tab:spk}, we evaluated the generated sentences with four captioning metrics that are BLEU-4\cite{papineni2002bleu}, ROUGE\cite{lin2004rouge}, CIDEr\cite{vedantam2015cider} and SPICE\cite{anderson2016spice}. The CMS is compared with an ablated architecture that is trained separately from the CMT, named CMS-type1, and with the Speaker-follower \cite{fried2018speaker}, 

\begin{table}[t]
\normalsize
\caption{\small Evaluation of generated sentences given the unseen validation set
}
\label{tab:spk}
\centering
\begin{tabular}{l|cccc}
\hline
 \multicolumn{1}{l|}{\bf Method }&{\bf BLEU-4}& {\bf CIDEr} &{\bf ROUGE}&{\bf SPICE} \\
\hline
Speaker \cite{fried2018speaker}  &{\bf 13.5}&27.2&{\bf 33.6}& 19.2 \\
CMS-type1 &6.4&{\bf 27.9}&27.6&20.9 \\
(Ours) CMS  &5.3&23.6& 27.3&{\bf 21.9} 
\end{tabular} 
\end{table}   

Although our approach performed better than the two other baselines only for SPICE metric, our results are consistent with the study \cite{zhao2021evaluation} that claims that standard captioning metrics, except for SPICE, are ineffective for VLN tasks.

\subsection{Qualitative Results}


The qualitative results of the CMT and CMS are illustrated on Fig. \ref{fig:qualitative}. Each rows represents a sample instructions from the validation set, and is given as a sequence of actions. The first three samples samples show successfully predicted path from the CMT. The CMS also generated consistent instructions, which suggest that the relation between visual and linguistic features was correctly modeled. The last row illustrates an erroneous sample for both path prediction and instruction generation. Such a sample illustrates one of the challenges of the VLN task, as the ``ping pong table'', which is a landmark, was seen from far and  from several positions. Similarly several tables were in the seen and should be differentiated from each other. These challenges also affected the generated instruction.


\section{Conclusion}

In the context of the increasing demand for DSRs, we addressed  visual navigation from natural language instruction in home environments. The mainresults of the study are summarized as follows:
\begin{itemize}
\item We proposed the CrossMap Transformer, which sequentially generates actions to reach an instructed destination determined from linguistic and visual features. 
\item We proposed the double back-translation to improve the mapping between linguistic and actions features using a common structure between a CrossMap Speaker network, which translates the sequence of actions into an instruction, and the CrossMap Transformer.
\Update
\item We achieved comparable results as state-of-the-art approaches using LSTM decoders
\end{itemize}
\Done

In future work, we will combine our present achievements with linguistic explanation, allowing interaction with non-expert users in failure cases.



\section*{ACKNOWLEDGEMENT}
This work was partially supported by JSPS KAKENHI Grant Number 20H04269, JST CREST, JST Moonshot R\&D Grant Number JPMJMS2011, and NEDO.

\squeezeup

\bibliographystyle{IEEEtran}
\bibliography{reference}

\end{document}